\documentclass{article}

\usepackage{microtype}
\usepackage{graphicx}
\usepackage{subcaption}
\usepackage{booktabs} 
\usepackage{enumitem}
\usepackage{hyperref}

\usepackage[preprint]{icml2026}

\usepackage{amsmath}
\usepackage{amssymb}
\usepackage{mathtools}
\usepackage{amsthm}
\usepackage{amsmath, amssymb}
\usepackage{algorithm}
\usepackage{algpseudocode}

\usepackage[capitalize,noabbrev]{cleveref}
\usepackage{balance}
\usepackage{listings}
\usepackage{xcolor}
\theoremstyle{plain}
\newtheorem{theorem}{Theorem}[section]

\newtheorem{corollary}[theorem]{Corollary}
\theoremstyle{definition}
\newtheorem{definition}[theorem]{Definition}
\newtheorem{assumption}[theorem]{Assumption}
\theoremstyle{remark}
\newtheorem{remark}[theorem]{Remark}

\usepackage[textsize=tiny]{todonotes}

\icmltitlerunning{MePoly: Max Entropy Polynomial Policy Optimization}

\begin{document}

\twocolumn[
  \icmltitle{MePoly: Max Entropy Polynomial Policy Optimization}



  \icmlsetsymbol{equal}{*}

  \begin{icmlauthorlist}
    \icmlauthor{Hang Liu}{umich}
    \icmlauthor{Sangli Teng}{berkeley}
    \icmlauthor{Maani Ghaffari}{umich}

  \end{icmlauthorlist}

  \icmlaffiliation{umich}{University of Michigan, Ann Arbor}
  \icmlaffiliation{berkeley}{University of California, Berkeley}

  \icmlcorrespondingauthor{Hang Liu}{hangliu@umich.edu}
  \icmlcorrespondingauthor{Sangli Teng}{sangliteng@berkeley.edu}
  \icmlcorrespondingauthor{Maani Ghaffari}{maanigj@umich.edu}

  \vskip 0.3in
]



\printAffiliationsAndNotice{}  

\begin{abstract}

Stochastic Optimal Control provides a unified mathematical framework for solving complex decision-making problems, encompassing paradigms such as maximum entropy reinforcement learning(RL) and imitation learning(IL). However, conventional parametric policies often struggle to represent the multi-modality of the solutions. Though diffusion-based policies are aimed at recovering the multi-modality, they lack an explicit probability density, which complicates policy-gradient optimization. To bridge this gap, we propose MePoly, a novel policy parameterization based on polynomial energy-based models. MePoly provides an explicit, tractable probability density, enabling exact entropy maximization. Theoretically, we ground our method in the classical moment problem, leveraging the universal approximation capabilities for arbitrary distributions. Empirically, we demonstrate that MePoly effectively captures complex non-convex manifolds and outperforms baselines in performance across diverse benchmarks\footnote{Code:\url{https://github.com/UMich-CURLY/MePoly}}.

\vspace{-2em}
\end{abstract}

\section{Introduction}

Many stochastic optimal control problems can be cast as learning a stochastic policy over continuous decisions.
This perspective naturally raises two foundational questions: \emph{what objective should we optimize}, and \emph{what policy family should we optimize over}. 
In the remainder of this section, we briefly review (i) the maximum-entropy (MaxEnt) view of policy optimization, and (ii) multimodal policy parameterizations that improve expressivity beyond unimodal Gaussians.
This background will clarify the practical tension between tractable information-theoretic objectives and expressive policy classes, which motivates the method developed in this work.

\begin{figure}
    \centering
    \includegraphics[width=1\linewidth]{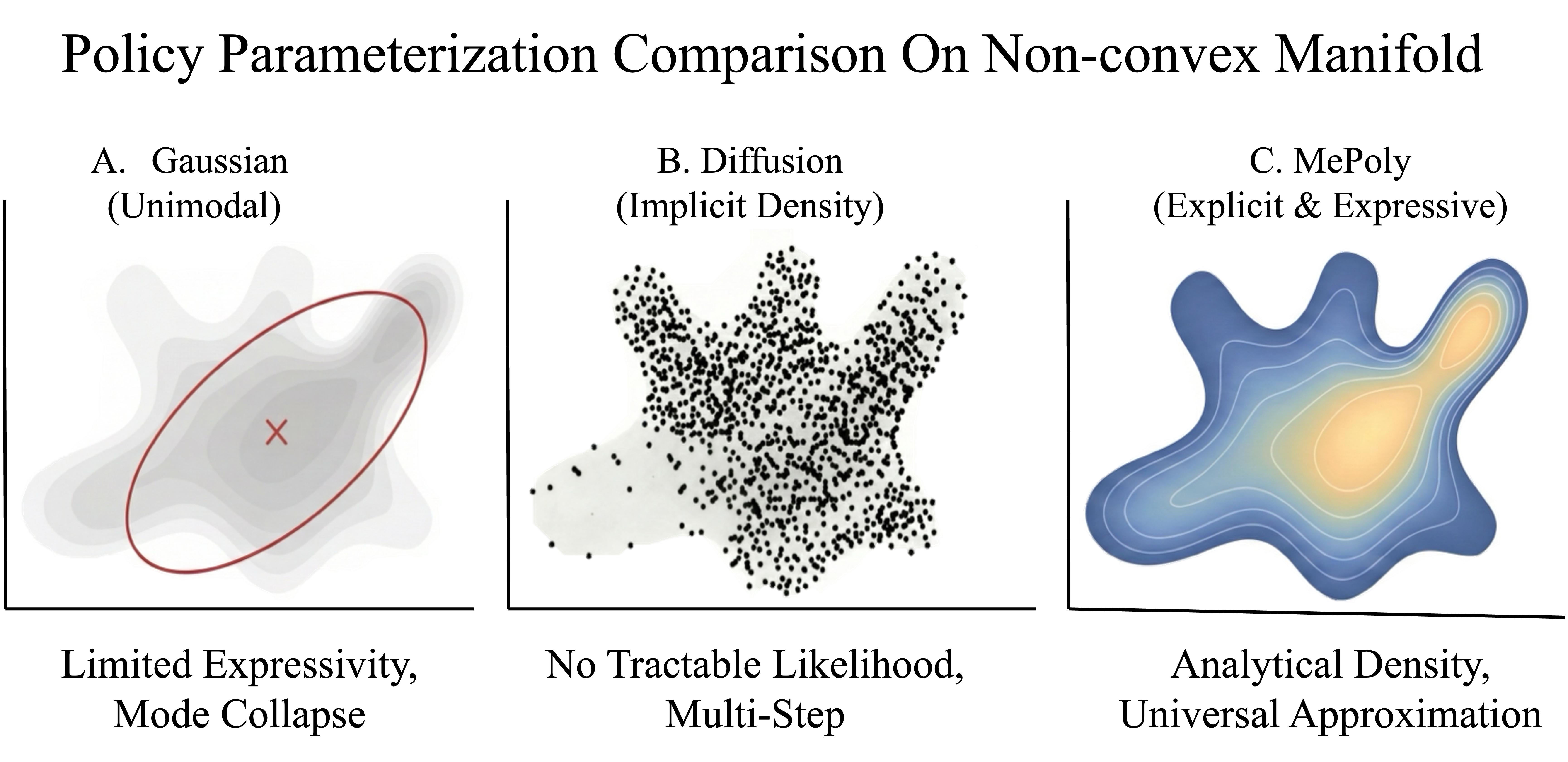}
    \caption{Conceptual comparison of policy parameterizations on a non-convex action manifold. (A) A unimodal Gaussian concentrates mass around a single mean, leading to limited expressivity and mode collapse. (B) Diffusion can represent complex supports via iterative sampling but lacks a tractable likelihood and requires multi-step generation. (C) MePoly yields an expressive, explicit density that can conform to the manifold while retaining tractable log-probabilities and entropies for learning.}
    \label{fig:cover}
    \vspace{-2em}
\end{figure}

 \subsection{Max Entropy Policy Optimization}

In information theory, the Shannon entropy of a random variable quantifies the expected information of its outcomes. The MaxEnt principle advocates selecting, among all distributions consistent with given constraints, the one with the largest entropy, thereby remaining maximally non-committal beyond what the constraints imply~\cite{jaynes1957information}.

In stochastic optimal control, the same principle leads naturally to soft optimality: instead of committing to a single optimal solution, one induces a distribution that assigns higher probability to higher-return solutions while retaining uncertainty~\cite{todorov2006linearly, kappen2005path}, which yields soft Bellman backups and a Boltzmann-form policy~\cite{ziebart2008maximum, fox2015taming, haarnoja2017reinforcement},

This perspective connects classical information-theoretic control, planning, and KL-regularized formulations. In path-integral control and model predictive path integral (MPPI), one samples trajectory rollouts under control perturbations and aggregates them using Gibbs weights, producing a soft update that accounts for multiple near-optimal solutions~\cite{kappen2005path,williams2016aggressive, williams2017information}. Similarly, cross-entropy model predictive control (MPC) updates a sampling distribution by minimizing a KL (cross-entropy) objective toward an elite set of rollouts, yielding an information-theoretic optimizer for trajectory planning~\cite{de2005tutorial}.

In modern reinforcement learning (RL) methods,  MaxEnt yields soft Bellman backups and a Boltzmann-form policy~\cite{ziebart2008maximum, fox2015taming, haarnoja2017reinforcement, haarnoja2018soft}, such as Soft Actor-Critic and Soft Q-learning. This MaxEnt RL becomes a paradigm for solving continuous control with better exploration~\cite{schulman2015trust, schulman2017proximal}. From a distributional viewpoint, both reinforcement learning and imitation learning (IL) can be cast as matching \mbox{$\pi(\cdot\mid s)$} to a target distribution: in RL, the target is \mbox{$\pi(a\mid s) \propto \exp\!\big(Q(s,a)\big)$}, while in imitation learning it is provided by expert demonstrations \mbox{$\pi(a\mid s) \propto \pi^{\text{expert}}(a\mid s)$}. However, the commonly used Gaussian parameterization limits the expressiveness and information of the policy.

\subsection{Multimodal Parameterization}
Recent energy-based and generative policy parameterizations, including diffusion models, flow-matching, and latent-conditioned generators~\cite{huang2023reparameterized}, have shown strong empirical performance for modeling complex action distributions. They have been applied successfully to behavior cloning~\cite{chi2024diffusionpolicy}, post-training refinement~\cite{ren2024diffusion}, online reinforcement learning~\cite{li2024learning,mcallister2025flow}, and offline reinforcement learning~\cite{wang2022diffusion}. 
However, these policies are typically trained with variational surrogate objectives (e.g., ELBO-style proxies), since exact maximum-likelihood is generally intractable; the resulting mismatch with RL objectives can complicate optimization and has been associated with mode collapse in online settings~\cite{li2024learning}. 
Classical multimodal alternatives such as Gaussian mixture policies~\cite{ren2021probabilistic} and Beta distributions~\cite{chou2017improving} improve expressivity over unimodal Gaussians but still struggle to represent highly nontrivial geometries (e.g., the lemniscate distributions in Sec.~\ref{exp-manifold}). 
Discretization-based policies~\cite{tang2020discretizing} also enable multimodality, but constructing joint distributions scales poorly for high-dimensional continuous action spaces.

These limitations point to a missing sweet spot: for continuous control with non-convex and multi-modal solution landscapes, we would like a policy class that is \emph{as expressive as modern generative parameterizations}, yet remains \emph{compatible with standard policy optimization} by providing an explicit density with tractable $\log \pi(a\mid s)$ and entropy/KL.
In particular, many MaxEnt and KL-regularized objectives rely on these quantities directly for stable updates; when they are unavailable or only accessible through surrogates, optimization can become fragile in practice.

To this end, we introduce \textbf{MePoly}, an expressive \emph{explicit} policy family based on a maximum-entropy construction with polynomial moment features.
MePoly yields a flexible exponential-family density on a compact support, allowing it to conform to highly nontrivial geometries while retaining tractable log probability and entropy.
This makes MePoly a drop-in replacement for common continuous-control objectives while substantially improving expressivity over previous methods (Figure.~\ref{fig:cover}). We summarize our main contributions as follows:

\begin{enumerate}[leftmargin=*,
    topsep=0pt, partopsep=0pt,
    itemsep=0.2em, parsep=0pt]
    \item \textbf{Universally-approximating polynomial MaxEnt distributions}: We ground MePoly in the classical moment problem and the maximum-entropy principle, showing that the MaxEnt polynomial exponential family forms a universally expressive policy class capable of approximating arbitrary target distributions.
    \item \textbf{Tractable likelihood and entropy}: MePoly admits an explicit density with tractable log-probabilities and entropy, computed via the log-partition function on compact support, enabling seamless integration with MaxEnt and KL-regularized objectives.
    \item \textbf{Efficient sampling and empirical evaluation}: We provide practical sampling mechanisms compatible with policy-gradient learning, allowing MePoly to be used as a stochastic policy in modern RL and IL pipelines. We demonstrate that MePoly improves expressivity and mitigates mode collapse.
\end{enumerate}

\begin{figure*}[h]
    \centering
    \includegraphics[width=0.8\linewidth]{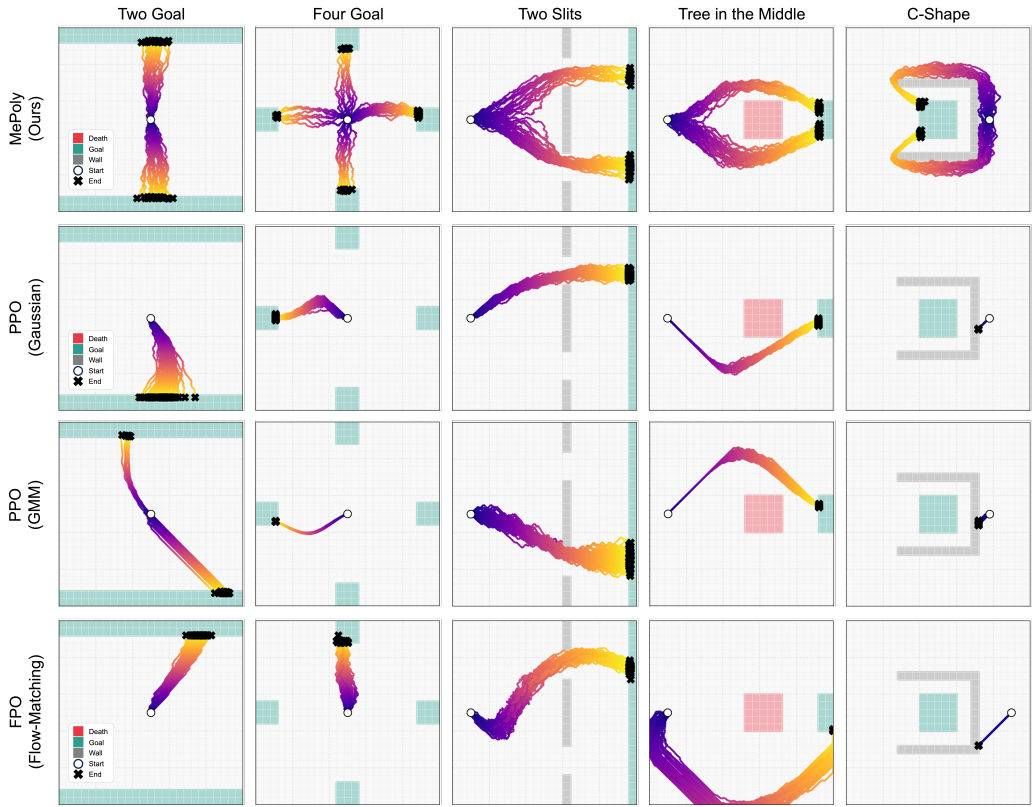}
    \caption{Trajectory samples on multi-modal navigation tasks: Each column is a different environment. Each row corresponds to various methods: MePoly (ours), PPO (Gaussian), PPO (Gaussian Mixture Model), and FPO (Flow-Matching). In every panel, we visualize multiple rollouts from the same start state (white circle) under identical environment layouts: walls/obstacles are shown in gray, goal regions in green, and death/unsafe regions in red (see legend). Trajectories are color-coded by time (purple → yellow), and black markers indicate terminal positions. MePoly consistently produces diverse, distinct feasible routes that cover multiple classes (e.g., different goals and passages through slits/around obstacles), whereas baselines often collapse to a single mode or fail to represent alternative valid solutions in highly non-convex environments.}
    \label{fig:fig-smoothworld-comp}
    \vspace{-2em}
\end{figure*}


\label{sec:exp_bandit}

\section{Preliminaries}
\label{sec:preliminaries}

We unify RL and IL within the framework of \textit{Maximum Entropy Policy Optimization}. This perspective treats the problem of policy learning as finding a distribution that minimizes a divergence to a target distribution—whether defined by a reward signal (in RL) or an expert dataset (in IL)—while having the least additional assumptions, i.e., maximizing the entropy.

\subsection{Problem Formulation: MDPs}
\label{subsec:mdp}

We consider a standard Markov Decision Process (MDP) defined by the tuple $\mathcal{M} = (\mathcal{S}, \mathcal{A}, p, r, \gamma)$, where $\mathcal{S} \subseteq \mathbb{R}^{n_s}$ is the state space, $\mathcal{A} \subseteq \mathbb{R}^{n_a}$ is the continuous action space, and $p(s_{t+1} | s_t, a_t)$ denotes the transition probability density. The reward function is defined as $r: \mathcal{S} \times \mathcal{A} \to \mathbb{R}$, and $\gamma \in [0, 1)$ is the discount factor. While we primarily formulate our method in the fully observable MDP setting for clarity, in partially observable scenarios (POMDPs), we denote the accessible information as observations $o \in \mathcal{O}$ and substitute the state $s$ with the observation $o$ without loss of generality.

\subsection{Maximum Entropy Framework for RL and IL}
\label{subsec:maxent_framework}

In the Maximum Entropy framework~\cite{ziebart2008maximum, haarnoja2017reinforcement}, we seek a stochastic policy $\pi(\cdot|s)$ that maximizes the expected cumulative objective while maintaining all possible solutions. The general objective is given by:
\begin{equation}
    \mathcal{J}_{\text{MaxEnt}}(\pi) = \mathbb{E}_{\pi} \left[ \sum_{t=0}^{\infty} \gamma^t \left( \mathcal{U}(s_t, a_t) + \alpha \mathcal{H}(\pi(\cdot|s_t)) \right) \right],
    \label{eq:maxent_general}
\end{equation}

where $\mathcal{U}(s_t, a_t)$ is a general utility term, $\mathcal{H}(\pi(\cdot|s_t)) = - \mathbb{E}_{a \sim \pi} [\log \pi(a|s_t)]$ is the entropy, and $\alpha$ is the coefficient controlling the stochasticity. This formulation unifies RL and IL by defining the utility term $\mathcal{U}$ and the underlying optimization procedure.

\paragraph{MaxEnt Reinforcement Learning.}
In the RL setting, the utility is defined as the extrinsic reward, $\mathcal{U}(s, a) := r(s, a)$. The optimal policy under this objective takes the form of a Boltzmann distribution~\cite{haarnoja2018soft}:
\begin{equation}
    \pi_{\phi}(a\mid s) \propto \exp\!\big(Q(s,a)\big),
    \label{eq:boltzmann_policy}
\end{equation}
where $Q(s,a)$, the action-value function, satisfies the soft Bellman equation. Our goal is to project this energy-based distribution onto a tractable parametric family $\pi_\phi$. This is commonly solved by maximizing the soft value function.

\paragraph{MaxEnt Imitation Learning.}
In the IL setting, given the expert demonstrations datasets $\mathcal{D}_E = \{(s_i, a_i)\}$, the goal is to learn a policy that matches the expert's distribution. This can be viewed as maximizing the log-likelihood of the expert actions, which is equivalent to minimizing the forward KL-divergence between the expert distribution and the policy:
\begin{equation}
    \max_\phi \mathbb{E}_{(s, a) \sim \mathcal{D}_E} [\log \pi_\phi(a|s)].
    \label{eq:bc_mle}
\end{equation}
In the context of energy-based models~\cite{florence2022implicit}, this corresponds to minimizing the energy of expert samples. If we parameterize the policy as an energy-based model $\pi_\phi(a|s) \propto \exp(E_\phi(s,a))$, behavior cloning (BC) becomes a direct maximum likelihood estimation (MLE) problem on the energy landscape.

\paragraph{Unified View.}
Both paradigms require the policy to represent complex, potentially multi-modal distributions defined by an energy function, either learned from rewards (Q-function) or directly from data (negative energy). The limitations of classical parameterizations in capturing these energies motivate our proposed \textbf{Polynomial Distribution Policy}, which provides a flexible, analytical, and expressive energy-based parameterization for both RL and IL.

\section{Polynomial Distribution Policy}

\subsection{Polynomial Distribution}
\label{subsec:poly_system}
We formulate the stochastic policy $\pi_\phi(a|s)$ as a conditional Energy-Based Model (EBM) \cite{lecun2006tutorial, florence2022implicit}. Unlike explicit policies that directly output action primitives (e.g., mean and variance of a Gaussian), or diffusion policies that learn a gradient field for iterative denoising \cite{chi2024diffusionpolicy}, our method explicitly structures the energy landscape using a \textbf{Polynomial}.

To formally establish the polynomial system and its associated distribution, we adopt the standard algebraic notation \cite{parrilo2000structured}. Let $\mathbb{R}[a]$ denote the ring of polynomials with real coefficients over the action variables $a := [a_1, \dots, a_{n_a}]^\top \in \mathbb{R}^{n_a}$. For notational brevity, we unify the learnable parameters as $\phi$.

\paragraph{Monomial Basis Definition.}
We define $\mathbb{N}$ as the set of non-negative integers $\{0, 1, \dots\}$. Given an order $K$, the set of valid exponent vectors $\mathbb{N}_K^{n_a}$ describes all possible combinations of degrees summing up to at most $K$:
\begin{equation}
    \mathbb{N}_K^{n_a} := \left\{ \alpha \in \mathbb{N}^{n_a} \;\middle|\; \sum_{i=1}^{n_a} \alpha_i \le K \right\}.
\end{equation}
A monomial is defined as $a^\alpha := \prod_{i=1}^{n_a} a_i^{\alpha_i}$. The total number of such basis functions corresponds to the dimension of the polynomial space, given by the binomial coefficient $M := \binom{n_a + K}{n_a}$.

\paragraph{Polynomial Energy and Joint Distribution.}
We define the feature vector $T(a): \mathbb{R}^{n_a} \to \mathbb{R}^M$ by defining a canonical ordering of all monomials in the set $\mathbb{N}_K^{n_a}$:
\begin{equation}
    T(a) = \big[ a^{\alpha_1}, a^{\alpha_2}, \dots, a^{\alpha_M} \big]^\top, \quad \text{indexed by } \alpha \in \mathbb{N}_K^{n_a}.
\end{equation}

Then we could know this polynomial action system $\langle \lambda_\phi(s), T(a) \rangle$, where $\lambda$ is the natural parameters predicted by a neural network. 
For example, consider a concrete case with a 3-dimensional action space with a 3rd-order polynomial basis, the energy expansion explicitly captures up to ternary dependencies:

\begin{equation}
\begin{split}
    \langle \lambda_\phi(s), T(a) \rangle &= 
    \lambda_0 \\
    &\quad + \lambda_1 a_1 + \lambda_2 a_2 + \lambda_3 a_3 \\
    &\quad + \lambda_{11} a_1^2 + \lambda_{22} a_2^2 + \lambda_{33} a_3^2 \\
    &\quad + \lambda_{12} a_1 a_2 + \lambda_{13} a_1 a_3 + \lambda_{23} a_2 a_3 \\
    &\quad + \lambda_{111} a_1^3 + \lambda_{222} a_2^3 + \lambda_{333} a_3^3 \\
    &\quad + \lambda_{112} a_1^2 a_2 + \lambda_{113} a_1^2 a_3 + \lambda_{122} a_1 a_2^2 \\
    &\quad + \lambda_{133} a_1 a_3^2 + \lambda_{223} a_2^2 a_3 + \lambda_{233} a_2 a_3^2 \\
    &\quad + \lambda_{123} a_1 a_2 a_3.
\end{split}
\label{eq:poly_expansion_full}
\end{equation}

This formulation naturally constructs a joint distribution $\pi_\phi(a|s) \propto  \langle \lambda_\phi(s), T(a) \rangle $ through the inclusion of cross-terms. Our basis set $\mathbb{N}_K^{n_a}$ explicitly includes mixed exponents (where multiple $\alpha_i > 0$, such as $a_i a_j$), thereby capturing the correlations and coupling between different action dimensions.

\paragraph{Legendre Polynomials.} 
While Eq.~\ref{eq:poly_expansion_full} uses monomials for exposition, implementing MePoly with standard monomials often yields degenerate, spectrally overlapping features: derivatives of $a^k$ vanish near the origin, reducing sensitivity to local, high-frequency changes and complicating optimization. To improve stability, we instead use an orthogonal Legendre basis~\cite{szeg1939orthogonal}. Its $L^2$ orthogonality decouples the natural parameters $\lambda_\phi(s)$ so that different orders capture distinct frequency components, acting as a spectral regularizer that produces smoother energy landscapes and improved robustness, consistent with our ablations Sec.\ref{exp-manifold}.

\paragraph{Energy-Based Parameterization.}
The policy distribution is defined by the Gibbs-Boltzmann distribution associated with an energy function $E_\phi(s, a)$:
\begin{equation}
    \pi_\phi(a|s) = \frac{\exp\left( -E_\phi(s, a) \right)}{\int \exp\left( -E_\phi(s, a) \right)},
    \label{eq:ebm_policy}
\end{equation}
The energy function $E_\phi(s, a)$ of the polynomial system is $-\langle \lambda_\phi(s), T(a) \rangle$:
\begin{equation}
    \pi_\phi(a|s) = \exp\left( \underbrace{\langle \lambda_\phi(s), T(a) \rangle}_{-E_\phi(s, a)} - A(\lambda_\phi(s)) \right),
\end{equation}
where $A(\lambda_\phi(s))  = \log\int \exp(\langle \lambda_\phi(s), T(a) \rangle) da$ is the log-partition function. By normalizing actions to a compact domain $a \in [-1, 1]^{n_a}$, the integral for $A(\cdot)$ remains finite and tractable (Details in Sec.~\ref{method-integral}):
\begin{equation}
    \log Z = \log \int_{-1}^{1} \exp(\langle \lambda_\phi(s), T(a) \rangle) da.
\end{equation}


\subsection{Why the Polynomial Distribution?}
\label{subsec:why_poly}

The choice of parameterizing the policy $\pi_\phi(a|s)$ via a polynomial energy function is grounded in the classical moment problem. The polynomial distribution offers a rigorous theoretical guarantee of universal approximation for any valid belief distribution on a compact domain.

We formalize this advantage through the lens of \textit{Moment-Constrained Max-Entropy Distributions} (MED). First, we define the class of distributions we aim to capture.

\begin{definition}[Moment-Determinate Distribution \cite{schmudgen2017moment}]
\label{def:moment_determinate}
A probability distribution $p(x)$ is said to be moment-determinate if it is uniquely determined by the sequence of all its moments $\mathbb{E}_{x\sim p}[x^\alpha]$ for $\alpha \in \mathbb{N}^n$.
\end{definition}

In the context of robotic control, the optimal policy (expert behavior) often exhibits complex multi-modal structures. The following theorem establishes that our polynomial policy can asymptotically recover \textbf{any such distribution}.

For many control applications, it is natural to assume a bounded action space: actuation limits, kinematic/dynamic feasibility, and safety constraints preclude infinitely large control commands. We therefore assume our policy distribution is supported on a compact set.

\begin{assumption}[Compact Support]
\label{assump:compact}
We assume the action space is bounded by box constraints, i.e., $a \in \mathcal{K} := [-a_{\max}, a_{\max}]^{n_a}$ for some $a_{\max} > 0$.
For notational convenience, we will rescale actions and henceforth take $a_{\max}=1$, i.e., $\mathcal{K}=[-1,1]^{n_a}$.
\end{assumption}

This mild assumption yields an important theoretical consequence: the induced policy distribution is moment-determinate, which is a key condition for the asymptotic convergence results below.

\begin{remark}
    By Proposition 12.17 in \cite{schmudgen2017moment}, a distribution with compact support is always moment determinate, thus ensuring the convergence of our polynomial policy to a unique solution when increasing $K$. 
\end{remark}

\begin{theorem}[Asymptotic Approximation \cite{borwein1991convergence, teng2025max}]
\label{thm:asymptotic}
 Consider the Max-Entropy distribution $p_K^*(a)$ constrained by moments up to order $K$. As the polynomial order $K \to \infty$, there exist unique limit $p_{\infty}$ where the distribution $p_K^*(a)$ converges to in the $L_1$ norm:
\begin{equation}
    \lim_{K \to \infty} \| p_K^*(a) - p_{\infty}(a) \|_1 = 0.
\end{equation}
\end{theorem}




\begin{corollary}[Universal Representational Capacity]
\label{cor:universal_approx}
A direct consequence of Theorem~\ref{thm:asymptotic} is that the polynomial parameterization serves as a universal approximator for stochastic control. It is capable of modeling arbitrary distributions on the compact support, covering both multi-modal expert policies in imitation learning and complex energy landscapes induced by action-value functions (i.e., $E(s,a) \approx -Q(s,a)$) in Reinforcement Learning. 
\end{corollary}


With the above theoretical support, a key question would be \emph{what's the advantage over the generative parameterizations?}

Compared to generative parameterizations (e.g., diffusion policies), our polynomial formulation offers three practical advantages for MaxEnt RL and real-time control. 
First, diffusion models are score-based, learning $\nabla \log \pi$ rather than an explicit density, so evaluating the exact log-likelihood (and thus the MaxEnt entropy term $\mathcal{H}(\pi)=-\mathbb{E}[\log \pi]$) typically requires costly probability-flow ODE rollouts, whereas MePoly admits a \textbf{tractable analytical log-density}. 
Second, diffusion models are highly expressive but often lack an explicit low-frequency inductive bias, making them prone to fitting high-frequency noise in limited-data regimes; in contrast, the polynomial order $K$ naturally serves as a spectral regularizer, constraining the learned energy landscape to \textbf{smooth, low-frequency components}. 
Third, diffusion inference relies on iterative denoising and incurs substantial latency, while MePoly reduces action selection to a \textbf{single-step energy minimization}, which can be interpreted as solving a parametric optimal control problem (OCP), enabling low-latency decisions suitable for high-frequency control loops.

\subsubsection{Tractable Implementation and Numerical Stability}
\label{method-integral}
We implement MePoly with a GPU-parallel numerical quadrature scheme, where a deep network predicts the natural parameters $\lambda_\phi(s)$. 
To avoid intractable analytic integration of the log-partition function $A(\lambda)$, we evaluate it via numerical quadrature on a fixed grid over the compact support $[-1,1]$, which is numerically stable and preserves differentiability of the log-likelihood for gradient-based optimization. 
For sampling, we form a discrete CDF from the quadrature grid masses and draw samples using a vectorized binary search, effectively treating the continuous density as a high-resolution discrete distribution at inference time to enable high-throughput sampling without iterative latency. 
We use deterministic tensor-product quadrature for stability and GPU efficiency, but the formulation also supports stochastic estimators (Monte Carlo, importance sampling) and more scalable deterministic cubature rules (e.g., sparse-grid/Smolyak quadrature) as action dimension increases.

\section{Learning Objectives}
\label{subsec:learning}

Our framework adapts naturally to both Reinforcement Learning and Imitation Learning, albeit with distinct training strategies to accommodate the latent structure.

\subsection{MePoly for Reinforcement Learning (RL).}
In the RL problem, where we consider online setting,the agent interacts with the environment to maximize returns. Consequently, we directly parameterize the policy head using the proposed polynomial distribution applied to the action space ($a \in \mathbb{R}^{n_a}$). To optimize the policy via PPO~\cite{schulman2017proximal}, we first define the standard clipped surrogate objective as:
\begin{equation}
    \mathcal{L}^{\text{CLIP}}_t(\phi) = \min \left( r_t(\phi) \hat{A}_t, \text{clip}(r_t(\phi), 1-\epsilon, 1+\epsilon)\hat{A}_t \right).
\end{equation}
Subsequently, the final objective function incorporating entropy regularization is formulated as:
\begin{equation}
    \mathcal{L}_{\text{RL}} = \mathbb{E}_{t} \left[ \mathcal{L}^{\text{CLIP}}_t(\phi) + \beta \mathcal{H}(\pi_\phi(\cdot|s_t)) \right].
\end{equation}
Crucially, the exact entropy term $\mathcal{H}$ derived from our polynomial system provides superior exploration signals compared to conventional parameterizations.

\subsection{MePoly for Imitation Learning (IL).}
Current state-of-the-art methods in imitation learning, such as ACT \cite{zhao2023learning, zhang2025action} and Diffusion Policy \cite{chi2024diffusionpolicy}, demonstrate that \textit{action chunking}, predicting a sequence of future actions $a_{t:t+H}$ rather than a single step, is crucial for temporal consistency and smooth control. However, naive action chunking necessitates modeling a joint distribution over a high-dimensional space $\mathbb{R}^{n_a \times H}$. As the prediction horizon $H$ grows, the curse of dimensionality makes learning the precise joint density computationally prohibitive and prone to overfitting.


\paragraph{The Manifold Hypothesis and Latent Actions.}
We hypothesize that the high-dimensional joint distribution over trajectories contains substantial redundancy and can be represented more compactly. In practice, feasible robotic motions are structured by \textbf{kinematic \& dynamic constraints} (e.g., joint limits, torque saturation, and inertia), \textbf{Lipschitz continuity} arising from smooth physical evolution across time steps~\cite{pan2025adonoisingdispellingmyths}, and \textbf{task structure} that concentrates goal-directed behaviors onto a sparse set of movement primitives or synergies~\cite{ijspeert2002movement}. As noted in \cite{losey2022learning}, a useful latent space for control should provide good conditioning, controllability, consistency, and scalability. Motivated by these observations, we posit that expert trajectories can be losslessly compressed into a low-dimensional latent manifold $\mathcal{Z}\subset\mathbb{R}^3$, where latent codes serve as global parameters that govern the motion.

\paragraph{Stage 1: Manifold Learning (Representation).} 
We first learn a compact latent representation for the high-dimensional action trajectory $\tau \in \mathbb{R}^{H \times n_a}$ using a VAE. The encoder $\mathcal{E}_\phi(\tau)$ outputs $(\mu_\phi(\tau), \sigma_\phi(\tau))$ and samples $z \in \mathbb{R}^{d_z}$ via reparameterization. The decoder $\mathcal{D}_\psi(z, s)$ reconstructs $\tau$ by combining a global intention code $z$ with the observation context $s$. Architecturally, we inject $z$ as a global modulator via Adaptive Layer Normalization (AdaLN), while $s$ is incorporated through cross-attention to provide local constraints. The training objective balances reconstruction and regularization, encouraging a well-structured latent manifold with compact support:
\begin{align}
\mathcal{L}_{\text{VAE}}
=
&\mathbb{E}_{\tau\sim\mathcal{D}}
\Big[
\|\tau-\mathcal{D}_\psi(z,s)\|_2^2 
+ \beta_{\text{dist}}\,\mathcal{L}_{\text{dist}}(\tau,z) \nonumber\\
&+ \beta_{\text{box}}\|\mathrm{ReLU}(|z|-\delta)\|_2^2
+ \beta_{\text{kl}}\,\mathrm{KL}\!\left(q_\phi(z\!\mid\!\tau)\,\|\,p(z)\right)
\Big],
\end{align}
where $p(z)$ is a fixed Gaussian prior and $\delta$ sets the latent compactness threshold.We use an MDS-style stress regularizer $\mathcal{L}_{\text{dist}}(\tau,z)$ that matches normalized pairwise distances between trajectories and their latent codes within a minibatch.

\paragraph{Stage 2: Policy Learning (Prior).} 
After learning the latent manifold, we train a prior policy $\pi_\phi(z\mid s)$ in the low-dimensional space $z\in[-1,1]^{d_z}$. Given an observation history $s$, a Prior Policy Network outputs the natural parameters $\lambda_\phi(s)$ of a polynomial exponential-family distribution. We use the encoder sample $z_{\text{gt}}\!\sim q_\phi(z\mid\tau)$ as the supervision target and maximize its likelihood with entropy regularization; in addition, we include a light reconstruction-consistency term and continue updating the decoder in a low learning rate to reduce the prior--decoder mismatch:
\begin{align}
\mathcal{L}_{\text{Poly}}
=
\mathbb{E}_{(s,\tau)\sim\mathcal{D}}
\Big[
&-\log \pi_\phi(z_{\text{gt}}\mid s)
-\alpha\,\mathcal{H}\!\left(\pi_\phi(\cdot\mid s)\right) \nonumber\\
&+\gamma\,\|\tau-\mathcal{D}_\psi(z_{\text{gt}},s)\|_2^2
\Big].
\end{align}
By performing density estimation in $\mathbb{R}^{d_z}$ instead of $\mathbb{R}^{H\times n_a}$, MePoly enables efficient multimodal sampling while avoiding the curse of dimensionality. Details are deferred to the Appendix.





\begin{figure*}[t]
    \centering
    \includegraphics[width=0.8\linewidth]{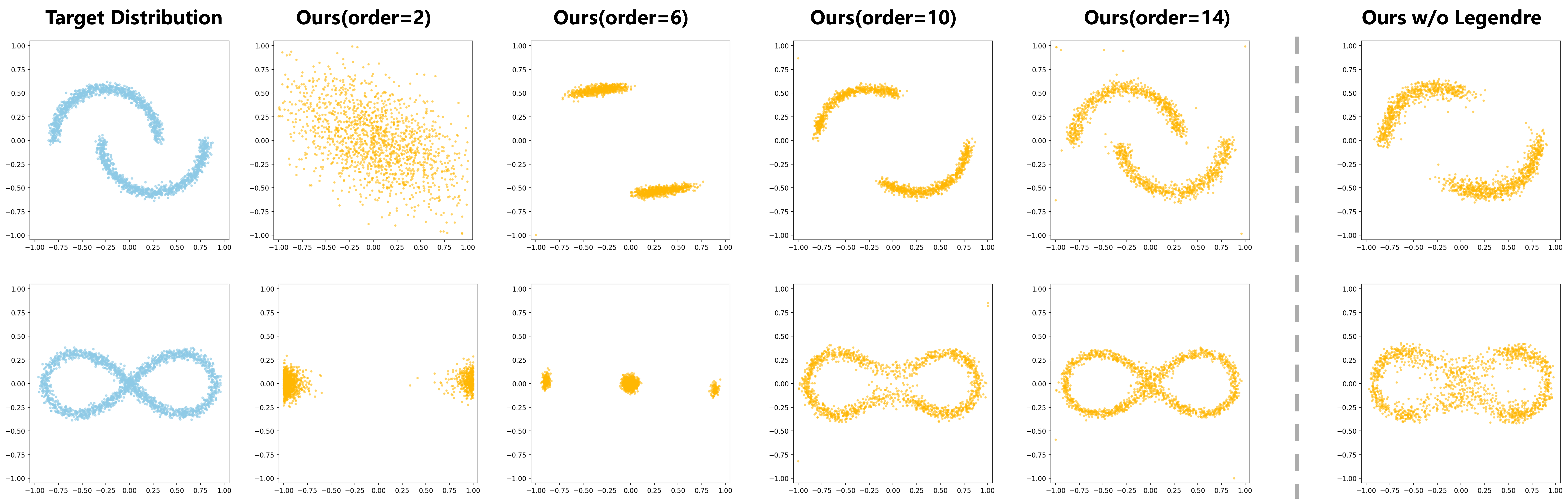}
    \caption{Myth of Mepoly: The LHS shows that Mepoly asymptotically approaches the sample distribution as the order increases, which validates Theorem~\ref{thm:asymptotic} and Corollary~\ref{cor:universal_approx}. The RHS replaces Legendre Polynomials with standard Polynomials, samples become noticeably blurrier and less faithful, indicating that orthogonal bases are critical for stable and accurate manifold approximation.}
    \label{fig:fig-bandit-abla}
    \vspace{-2em}
\end{figure*}

\begin{figure}[t]
    \centering
    \includegraphics[width=1\linewidth]{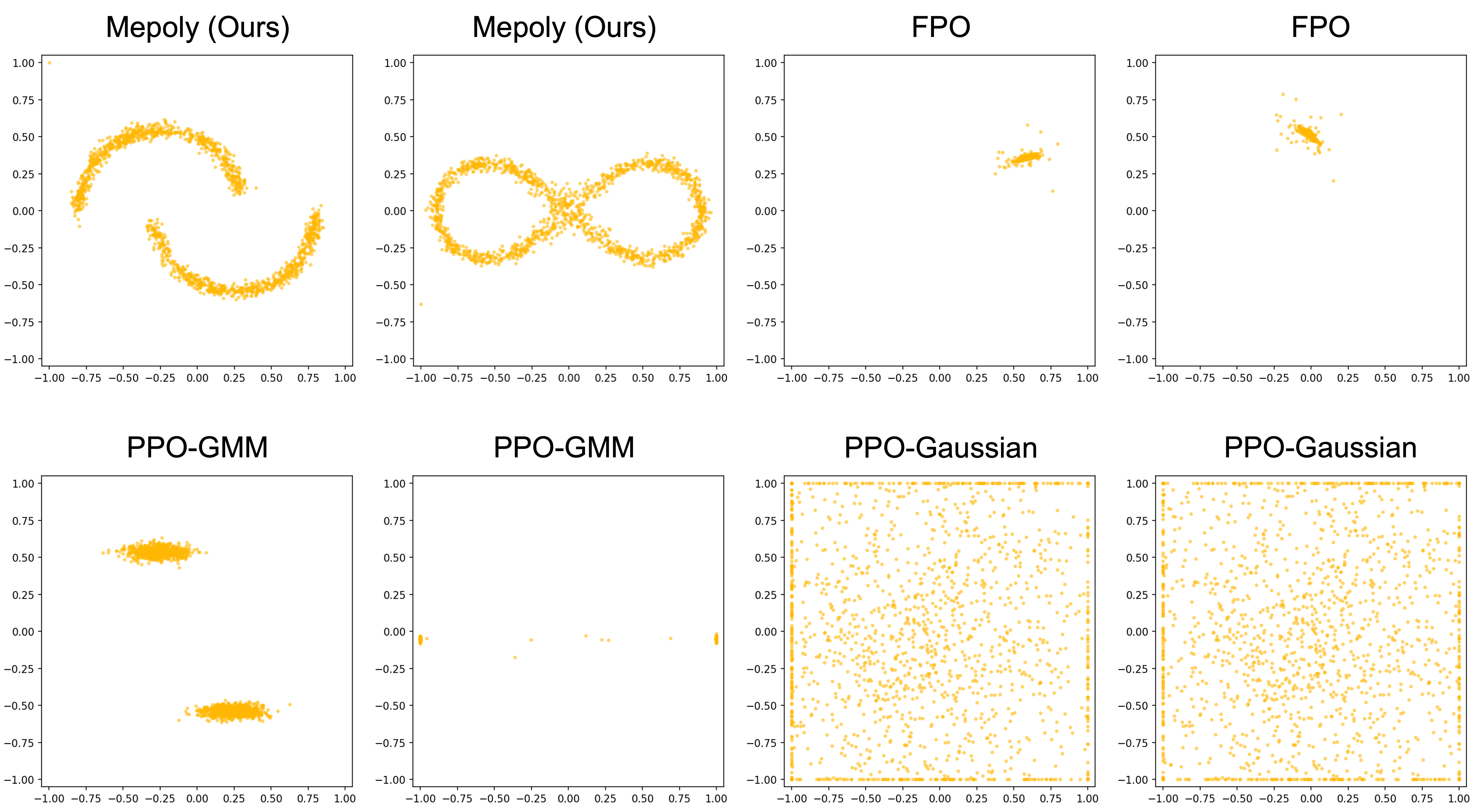}
    \caption{Sample quality comparison: MePoly (ours) generates samples that closely follow the underlying curved support and preserve the global topology. In contrast, the baselines fail to capture the non-convex and multi-modal structures, resulting in significant geometric mismatch and a collapse to unimodal or fragmented approximations.}
    \label{fig:fig-bandit-comp}
    \vspace{-2em}
\end{figure}

\section{Experiments}
Our experiments aim to answer the following questions: (1) Does our MePoly accurately capture a
multi-modal policy distribution in RL? (2) How complex a distribution can our MePoly capture? (3) Can our method also work in high-dimensional space, like a manipulator in imitation learning?

\paragraph{Baselines.} 
We evaluate MePoly against a diverse set of baselines tailored to each task domain. 
For RL problems, we compare our method with classical unimodal parameterizations (PPO-Gaussian)~\cite{schulman2017proximal} and their Gaussian mixture model extensions (PPO-GMM). 
Additionally, we benchmark against flow-matching-based policies (FPO), representing the state-of-the-art in generative policy gradients~\cite{mcallister2025flow}. 
For IL tasks, we primarily compare against Diffusion Policy (DP)~\cite{chi2024diffusionpolicy}, the current benchmark for expressive robot learning.

\paragraph{Scope and Integration.} 
To provide a unified perspective on the expressive power of polynomial policies, our analysis focuses specifically on the impact of the policy parameterization itself. 
Consequently, we do not evaluate orthogonal algorithmic improvements such as off-policy or offline RL variants, and we do not aim for a specific domain benchmark. 
It is important to note that MePoly is designed as a plug-and-play module, capable of being integrated into most existing policy-based optimization frameworks with minimal modification.

\subsection{Smooth World Navigation}
To validate the fundamental hypothesis that MePoly can capture complex, multi-modal action distributions where unimodal baselines fail, we introduce \textit{Smooth World}, inspired by ~\cite{mcallister2025flow}. This is a continuous 2D navigation environment designed to visualize the policy's energy landscape directly against physical constraints.

\textbf{Environment Setup.} The state space $\mathcal{S} \subseteq \mathbb{R}^2$ represents the agent's position normalized to $[-1, 1]$, and the action space $\mathcal{A} \subseteq \mathbb{R}^2$ represents continuous velocity vectors. The dynamics follow simplified kinematics: $p_{t+1} = p_t + v_t \cdot \Delta t$, subject to boundary and obstacle collisions. The reward function is sparse: the agent receives a positive reward $+R_{goal}$ upon entering a goal region, a penalty $-R_{death}$ for entering "death zones," and zero otherwise. This sparsity, combined with specific geometric arrangements, creates rigorous tests for exploration and policy expressivity.


    
    
    

\textbf{Result and Analysis.} 
Fig.~\ref{fig:fig-smoothworld-comp} reveals a clear difference in how policy parameterizations handle multimodal, non-convex action landscapes. PPO-Gaussian repeatedly collapses to a single mode: policy-gradient updates are greedy, and a unimodal Gaussian concentrates mass around one mean, amplifying whichever mode first attains slightly higher estimated advantage. Flow-matching baselines (FPO) are in principle multimodal, but in reward-sparse settings they often fail to represent sharp density transitions, partly because score-matching does not provide an explicit likelihood/entropy signal aligned with the MaxEnt objective.

In contrast, MePoly maintains probability mass over multiple disjoint manifolds. Its polynomial energy form yields an explicit (quadrature-evaluated) density and enables exact entropy maximization, allowing it to recover the full multimodal solution set rather than a single averaged behavior.

\subsection{Learning Complex Manifolds}
\label{exp-manifold}

In this experiment, we aim to disclose the myth of our Mepoly and demonstrate the extreme of our method. We design a \textit{Contextual Bandit} problem (single-step RL) where the agent must learn to match a static, complex target distribution solely through reward feedback.

\textbf{Environment Setup.} The environment consists of a single dummy state $s_0$. In each episode $t$, the agent takes a single action $a_t \in [-1, 1]^2$. The objective is to approximate a target geometry defined by a point sets $\mathcal{D} = \{x_i\}_{i=1}^N \sim \mathcal{P}_{target}$. We select two topologically challenging distributions, the lemniscate and the two moons.

\textbf{Manifold Reward.} The reward function serves as a proxy for the unnormalized density. We define the reward based on the distance to the nearest point on the target manifold:
\begin{equation}
    r(a) = \exp\left( -\frac{\min_{x \in \mathcal{D}} \|a - x\|^2}{2\sigma^2} \right)
\end{equation}
where $\sigma$ controls the width of the reward kernel (set to $0.05$ in our experiments). 

Under the Maximum Entropy objective $J(\pi) = \mathbb{E}[r(a)] + \alpha \mathcal{H}(\pi)$, the optimal policy is exactly the Boltzmann distribution $\pi^*(a) \propto \exp(r(a)/\alpha)$. Therefore, a successful agent must spread its probability mass uniformly across the entire \textit{Lemniscate} or \textit{Two Moons} structure, rather than collapsing to a single high-reward point.

\textbf{Result and Analysis.}

As shown in Fig.~\ref{fig:fig-bandit-abla}, we empirically validate Theorem~\ref{thm:asymptotic} and Corollary~\ref{cor:universal_approx}: increasing the polynomial order improves approximation to complex target distributions. Legendre bases are orthogonal and spectrally distinct across orders (e.g., a 4th-order basis exhibits a W''-like shape whereas the 2nd-order remains U''-shaped), in contrast to monomials that are structurally similar across degrees and thus redundant. This orthogonality yields a better-conditioned optimization landscape, keeping higher-order terms responsive to local geometry. As a result, MePoly captures sharp features such as moon cusps and narrow bottlenecks that are typically smoothed out by non-orthogonal parameterizations.

We further compare MePoly with strong baselines in Fig.~\ref{fig:fig-bandit-comp}. MePoly consistently achieves higher expressivity, largely because it provides exact log-density and entropy, which are critical for stable MaxEnt optimization. In contrast, PPO-GMM typically relies on surrogate or biased gradients (e.g., Gumbel-Softmax-style relaxations) due to non-differentiable mixture assignments, which can undermine multimodal learning. Likewise, FPO lacks exact log-probability/entropy signals, making generative policies more prone to policy-gradient greediness and mode collapse, whereas MePoly’s analytical density better preserves the diversity of the target manifold.

\subsection{ManiSkill}
\textbf{Environment Setup.} Our imitation learning (IL) experiments are conducted across four diverse tasks within the ManiSkill benchmark~\cite{mu2021maniskill}. We evaluate MePoly on a representative suite of manipulation tasks: (i) \textit{PushCube}, (ii) \textit{PickCube}, (iii) \textit{StackCube}, and (iv) \textit{PushT}. For both MePoly and Diffusion Policy, we control the model capacity by matching the total number of trainable parameters to 12M.

\textbf{Result and Analysis.}
Table~\ref{tab:combined_results} shows that MePoly performs comparably to Diffusion Policy (DP) on ManiSkill imitation learning benchmarks, with task-dependent gaps.
On \textbf{PushCube} and \textbf{StackCube}, MePoly achieves higher peak success rates, suggesting that an explicit low-dimensional latent prior can capture dominant demonstration modes without sacrificing stability.
On \textbf{PickCube} and \textbf{PushT}, DP slightly outperforms MePoly, while MePoly exhibits larger variance across random seeds.
Overall, MePoly matches DP on simpler tasks and also highlights remaining limitations.
A plausible cause is that the current objective for constructing the latent space is still imperfect that the latent manifold may be overly distorted.
Nevertheless, these results support the feasibility of performing imitation learning by controlling a compact action latent space in MePoly, and improving the construction of a well-behaved action latent space is an important direction for future work.

\begin{table}[t]
\centering
\caption{Comparative analysis of MePoly performance. \textbf{Top}: ManiSkill manipulation tasks demonstration. \textbf{Bottom}: Quantitative success rate comparison on ManiSkill benchmarks against the Diffusion Policy (DP). Results are aggregated over three random seeds; for each seed, we report the peak success rate achieved during training, and summarize performance as mean/std across seeds.}
\label{tab:combined_results}

\begin{center}
    \includegraphics[width=1\linewidth]{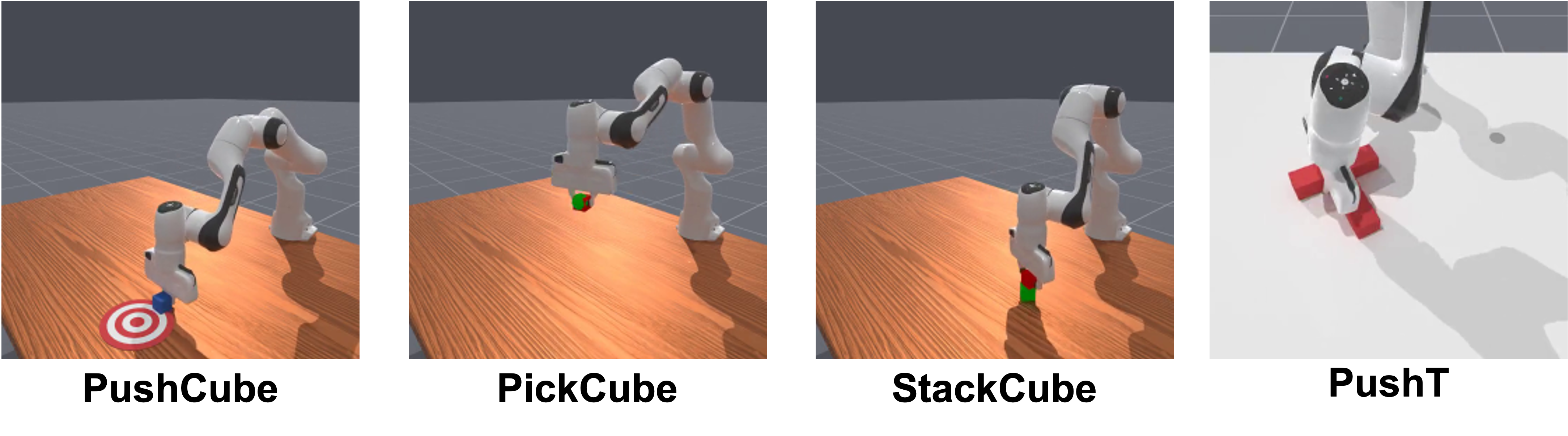}
\end{center}

{\scriptsize
\setlength{\tabcolsep}{8pt} 
\renewcommand{\arraystretch}{1.05} 
\begin{tabular}{@{}lcccc@{}}
\toprule
\textbf{Method} & \textbf{PushCube} & \textbf{PickCube} & \textbf{StackCube} & \textbf{PushT} \\ \midrule
MePoly      & 0.901/0.018 & 0.945/0.025 & 0.446/0.049 & 0.279/0.020 \\
DP          & 0.856/0.015 & 0.992/0.002 & 0.410/0.001 & 0.298/0.013 \\ \bottomrule
\end{tabular}
}
\end{table}

\section{Conclusion and Discussion}
We presented MePoly, a polynomial exponential-family policy that yields an \emph{explicit} and flexible density on compact support, enabling MaxEnt/KL-regularized optimization while capturing multi-modal, non-convex action distributions. Our results indicate that MePoly can mitigate mode collapse and remain competitive on ManiSkill imitation learning, suggesting that controlling a compact action latent space is a viable alternative to iterative generative policies. Key limitations are (i) the scalability of grid-based normalization and (ii) the sensitivity of imitation performance to the learned latent geometry. Future work will therefore focus on more scalable normalizers and improved representation learning objectives that produce better-behaved action latents, with the goal of strengthening robustness on long-horizon contact-rich tasks.

\newpage

\nocite{langley00}

{\small
\balance
\bibliography{example_paper}
\bibliographystyle{icml2026}
}

\newpage
\appendix
\onecolumn

\section{Implementation Details of the Two-Stage Training}\label{app:arch}

This section provides the implementation-level details omitted in the main text, including the exact network structures and the polynomial prior parameterization.

\paragraph{Data slicing and horizons.}
From each demonstration trajectory, we construct fixed-length training pairs by slicing with padding:
an observation history $s\in\mathbb{R}^{H_o\times n_s}$ and an action sequence $\tau\in\mathbb{R}^{H\times n_a}$,
where $H_o=\texttt{obs\_horizon}$ and $H=\texttt{pred\_horizon}$.
Padding repeats the boundary state/action to ensure each slice has the same length.

\paragraph{Trajectory VAE backbone.}
We use a Transformer-based VAE with latent dimension $d_z=3$ and embedding size $d=256$.
\begin{itemize}\setlength\itemsep{2pt}
    \item \textbf{Action encoder} $\mathcal{E}_\phi$:
    a linear embedding $\tau_t \mapsto \mathbb{R}^d$, plus a learned positional embedding;
    a TransformerEncoder with $3$ layers ($4$ heads, batch-first);
    mean pooling over time; two MLP heads for $\mu_\phi(\tau)\in\mathbb{R}^{d_z}$ and $\log\sigma_\phi^2(\tau)\in\mathbb{R}^{d_z}$.
    \item \textbf{Observation encoder}:
    a linear embedding for $s$ and a TransformerEncoder with $2$ layers ($4$ heads) producing a memory $M(s)\in\mathbb{R}^{H_o\times d}$.
    \item \textbf{Decoder} $\mathcal{D}_\psi$:
    a learned query positional embedding of length $H$;
    $4$ stacked decoder blocks (self-attention $\rightarrow$ cross-attention to $M(s)$ $\rightarrow$ FFN);
    a final linear head to predict $\hat{\tau}\in\mathbb{R}^{H\times n_a}$.
\end{itemize}

\paragraph{AdaLN + Cross-Attention conditioning.}
Each decoder block injects $z$ via Adaptive Layer Normalization (AdaLN): we apply LayerNorm without affine
and predict a per-channel scale/shift from $z$ (SiLU + Linear to $2d$). The AdaLN projection is initialized to zero,
so the decoder starts close to a standard Transformer and gradually learns to use $z$.
Observations enter through cross-attention using the observation memory $M(s)$ as keys/values.

\paragraph{Stage-1 objective (VAE).}
Training minimizes reconstruction error with regularization:
\begin{align}
\mathcal{L}_{\text{VAE}}
=
\mathbb{E}\Big[
&\|\tau-\mathcal{D}_\psi(z,s)\|_2^2
+ \beta_{\text{iso}}\mathcal{L}_{\text{iso}}(\tau,z)
+ \beta_{\text{box}}\|\mathrm{ReLU}(|z|-\delta)\|_2^2 \nonumber\\
&+ \beta_{\text{kl}}\mathrm{KL}\!\left(q_\phi(z\!\mid\!\tau)\,\|\,\mathcal{N}(0,\sigma_0^2 I)\right)
\Big],
\end{align}

$\mathcal{L}_{\text{iso}}$ matches pairwise distances between flattened trajectories and latents (MSE between normalized distance matrices),
discouraging manifold folding.

\paragraph{Polynomial prior distribution.}
We model $\pi_\theta(z\mid s)$ as a polynomial exponential-family distribution on $[-1,1]^{d_z}$:
\begin{align}
\pi_\theta(z\mid s) \propto 
\exp\!\Big(\langle \lambda_\theta(s),\,T(z)\rangle\Big),
\qquad
T(z) = \{z^{\alpha}\}_{|\alpha|_1\le K},
\end{align}
where $K=\texttt{poly\_order}$ and the feature set includes all monomials with total degree $\le K$.
The prior network is an MLP (hidden sizes $[256,256]$ with ReLU) mapping the flattened observation history to
$\lambda_\theta(s)\in\mathbb{R}^{|T|}$; we clip $\lambda$ elementwise for numerical stability.

\paragraph{Log-partition, entropy, and sampling.}
We approximate integrals on a fixed quadrature grid: a 1D grid in $[-1,1]$ with trapezoidal weights is expanded
to a Cartesian grid for $d_z\le 5$ (our setting uses $d_z=3$).
We precompute grid features $T(\bar{z})$ and evaluate
\begin{align}
\log Z(\lambda)=\log\sum_{\bar{z}\in\mathcal{G}}
\exp\!\Big(\langle \lambda, T(\bar{z})\rangle + \log w(\bar{z})\Big),
\end{align}
using \texttt{logsumexp} for numerical stability.
Entropy is computed from the induced discrete distribution on the grid.
Sampling draws a grid point by inverse-CDF on the discrete mass.

\paragraph{Stage-2 objective (Prior + continued decoder update).}
In the prior-learning stage, we obtain supervision by sampling
$z_{\text{gt}}\sim q_\phi(z\mid\tau)$ (encoder inference is performed without gradient).
We optimize NLL with entropy regularization, and add a light reconstruction-consistency term:
\begin{align}
\mathcal{L}_{\text{Poly}}
=
\mathbb{E}\Big[
&-\log \pi_\theta(z_{\text{gt}}\mid s)
-\alpha\,\mathcal{H}\!\left(\pi_\theta(\cdot\mid s)\right)
+\gamma\,\|\tau-\mathcal{D}_\psi(z_{\text{gt}},s)\|_2^2
\Big].
\end{align}
Implementation-wise, we update the prior network and \emph{continue updating the decoder}
(with a smaller learning rate). Concretely, we freeze the action encoder and latent heads,
and update the observation encoder and decoder modules (including the final action head).
We apply gradient clipping on the decoder parameters during this stage.

\paragraph{Optimization.}
We use AdamW throughout. Stage 1 updates the full VAE. Stage 2 uses two learning rates:
one for the prior network and a smaller one for the (continued) decoder update.
All remaining hyperparameters follow the configuration in our training script.

\section{Polynomial Joint Distribution Implementation Example}
\label{app:poly_joint_impl}

\begin{lstlisting}[language=Python, caption={Implementation example of the joint polynomial exponential-family distribution with quadrature-based normalization, entropy, and sampling.}, label={lst:poly_joint_dist}]
class PolynomialJointDistribution(nn.Module):
    """
    Joint polynomial exponential-family distribution on [-1, 1]^D.
    Uses a multi-dimensional quadrature grid for log-partition, entropy, and sampling.
    """

    def __init__(
        self,
        act_dim: int,
        order: int,
        grid_size: int = 64,
        lambda_clip: float = 5.0,
        action_eps: float = 1e-4,
        full_grid_max_dim: int = 3,
        stochastic_grid_size: int = 4096,
    ) -> None:
        super().__init__()
        self.act_dim = act_dim
        self.order = order
        self.grid_size = grid_size
        self.lambda_clip = lambda_clip
        self.action_eps = action_eps

        exponents = self._build_exponents(act_dim, order)
        self.num_features = exponents.shape[0]
        self.register_buffer("exponents", exponents)

        grid = torch.linspace(-1.0, 1.0, grid_size)
        weights = torch.ones_like(grid)
        weights[0] = weights[-1] = 0.5
        weights = weights * (2.0 / (grid_size - 1))  # trapezoidal rule scaling
        log_weights_1d = weights.clamp_min(1e-12).log()

        if act_dim <= full_grid_max_dim:
            grid_points, log_weights = self._build_full_grid(grid, log_weights_1d)
        else:
            grid_points, log_weights = self._build_stochastic_grid(
                grid, log_weights_1d, stochastic_grid_size
            )

        self.register_buffer("grid_points", grid_points)
        self.register_buffer("log_weights", log_weights)
        self.register_buffer("grid_features", self._monomials(grid_points))

    @staticmethod
    def _build_exponents(act_dim: int, order: int) -> torch.Tensor:
        exponents = [
            powers
            for powers in itertools.product(range(order + 1), repeat=act_dim)
            if sum(powers) <= order
        ]
        return torch.tensor(exponents, dtype=torch.long)

    def _build_full_grid(
        self, grid: torch.Tensor, log_weights_1d: torch.Tensor
    ) -> tuple[torch.Tensor, torch.Tensor]:
        meshes = torch.meshgrid(*[grid for _ in range(self.act_dim)], indexing="ij")
        grid_points = torch.stack(meshes, dim=-1).reshape(-1, self.act_dim)

        weight_meshes = torch.meshgrid(
            *[log_weights_1d for _ in range(self.act_dim)], indexing="ij"
        )
        log_weights = torch.stack(weight_meshes, dim=-1).sum(dim=-1).reshape(-1)
        return grid_points, log_weights

    def _build_stochastic_grid(
        self,
        grid: torch.Tensor,
        log_weights_1d: torch.Tensor,
        sample_size: int,
    ) -> tuple[torch.Tensor, torch.Tensor]:
        idx = torch.randint(0, grid.shape[0], (sample_size, self.act_dim))
        grid_points = grid[idx]
        log_weights = log_weights_1d[idx].sum(dim=-1)
        log_correction = self.act_dim * math.log(grid.shape[0]) - math.log(sample_size)
        log_weights = log_weights + log_correction
        return grid_points, log_weights

    def _stable_lambda(self, raw_lambda: torch.Tensor) -> torch.Tensor:
        return torch.clamp(raw_lambda, -self.lambda_clip, self.lambda_clip)

    def _logits_on_grid(self, lambda_params: torch.Tensor) -> torch.Tensor:
        return torch.matmul(lambda_params, self.grid_features.t())

    def log_partition(self, lambda_params: torch.Tensor) -> torch.Tensor:
        logits = self._logits_on_grid(lambda_params)
        return torch.logsumexp(logits + self.log_weights, dim=-1)

    def _monomials(self, x: torch.Tensor) -> torch.Tensor:
        base_shape = x.shape[:-1]
        legendre = x.new_zeros(*base_shape, self.act_dim, self.order + 1)
        legendre[..., 0] = 1.0
        if self.order >= 1:
            legendre[..., 1] = x
        for n in range(2, self.order + 1):
            n_float = float(n)
            legendre[..., n] = ((2 * n - 1) / n_float) * x * legendre[..., n - 1] - (
                (n - 1) / n_float
            ) * legendre[..., n - 2]

        orders = self.exponents.unsqueeze(0).expand(*base_shape, -1, -1).unsqueeze(-1)
        legendre = legendre.unsqueeze(-3).expand(*base_shape, self.num_features, self.act_dim, -1)
        selected = torch.gather(legendre, -1, orders).squeeze(-1)
        return selected.prod(dim=-1)

    def log_prob(self, lambda_params: torch.Tensor, action: torch.Tensor) -> torch.Tensor:
        features = self._monomials(action)
        log_z = self.log_partition(lambda_params)
        logits = (lambda_params * features).sum(dim=-1)
        return logits - log_z

    def entropy(self, lambda_params: torch.Tensor) -> torch.Tensor:
        logits = self._logits_on_grid(lambda_params)
        log_z = torch.logsumexp(logits + self.log_weights, dim=-1, keepdim=True)
        log_probs = logits - log_z
        probs = log_probs.exp()
        ent = -(probs * log_probs * self.log_weights.exp()).sum(dim=-1)
        return ent

    def sample(self, lambda_params: torch.Tensor):
        squeeze = False
        if lambda_params.dim() == 1:
            lambda_params = lambda_params.unsqueeze(0)
            squeeze = True

        logits = self._logits_on_grid(lambda_params)
        log_mass = logits + self.log_weights
        probs = torch.softmax(log_mass, dim=-1)
        cdf = probs.cumsum(dim=-1)
        u = torch.rand_like(cdf[..., :1])
        idx = torch.searchsorted(cdf, u, right=False).clamp(max=cdf.shape[-1] - 1)

        grid_expanded = self.grid_points.unsqueeze(0).expand(
            logits.shape[0], -1, -1
        )
        idx_expanded = idx.unsqueeze(-1).expand(-1, -1, self.act_dim)
        action = torch.gather(grid_expanded, 1, idx_expanded).squeeze(1)

        logprob = self.log_prob(lambda_params, action)
        if squeeze:
            return action.squeeze(0), logprob.squeeze(0)
        return action, logprob

    def expected_action(self, lambda_params: torch.Tensor) -> torch.Tensor:
        logits = self._logits_on_grid(lambda_params)
        log_mass = logits + self.log_weights
        probs = torch.softmax(log_mass, dim=-1)
        return (probs.unsqueeze(-1) * self.grid_points.unsqueeze(0)).sum(dim=-2)
\end{lstlisting}


\end{document}